\documentclass[sigconf]{acmart}
\usepackage{hyperref}       
\usepackage{url}            
\usepackage{booktabs}       
\usepackage{amsfonts}       
\usepackage{nicefrac}       
\usepackage{microtype}      
\usepackage{xcolor}         
\usepackage{graphicx}
\usepackage{multirow}
\usepackage{algorithm}
\usepackage{algorithmic}
\usepackage{amsmath}

\usepackage{amssymb}
\usepackage{xcolor}
\usepackage{mathtools}
\usepackage{enumitem}
\usepackage{amsthm}
\usepackage{makecell}
\usepackage{wrapfig}
\usepackage{multicol}
\usepackage{bm}

\definecolor{Hotel_Pink}{RGB}{232, 95, 91}
\definecolor{Kingdom_Yellow}{RGB}{255, 173, 88}
\definecolor{Aquatic_Blue}{RGB}{100, 161, 155}

\AtBeginDocument{%
  \providecommand\BibTeX{{%
    \normalfont B\kern-0.5em{\scshape i\kern-0.25em b}\kern-0.8em\TeX}}}

\setcopyright{acmcopyright}
\copyrightyear{2018}
\acmYear{2018}
\acmDOI{XXXXXXX.XXXXXXX}

\acmConference[Conference acronym 'XX]{Make sure to enter the correct conference title from your rights confirmation emai}{June 03--05,
  2018}{Woodstock, NY}
\acmPrice{15.00}
\acmISBN{978-1-4503-XXXX-X/18/06}

\begin{document}
\title{FGAD: Self-boosted Knowledge Distillation for An Effective Federated Graph Anomaly Detection Framework}
\author{Jinyu Cai}
\authornote{Both authors contributed equally to this research.}
\email{jinyucai@nus.edu.sg}
\affiliation{%
  \institution{Institute of Data Science, National University of Singapore}
  \country{Singapore}
  }

\author{Yunhe Zhang}
\authornotemark[1]
\email{zhangyhannie@163.com}
\affiliation{%
  \institution{School of Data Science, The Chinese University of HongKong (Shenzhen)}
  \country{China}
  }

\author{Zhoumin Lu}
\email{walker.zhoumin.lu@gmail.com}
\affiliation{%
  \institution{School of Computer Science, Northwest Polytechnical University}
  \country{China}
}

\author{Wenzhong Guo}
\email{guowenzhong@fzu.edu.cn}
\affiliation{%
  \institution{College of Computer and Data Science, Fuzhou University}
  \country{China}
}

\author{See-Kiong Ng}
\email{seekiong@nus.edu.sg}
\affiliation{%
  \institution{Institute of Data Science, National University of Singapore}
  \country{Singapore}
  }

\renewcommand{\shortauthors}{Cai, et al.}


\begin{abstract}
Graph anomaly detection (GAD) aims to identify anomalous graphs that significantly deviate from other ones, which has raised growing attention due to the broad existence and complexity of graph-structured data in many real-world scenarios. However, existing GAD methods usually execute with centralized training, which may lead to privacy leakage risk in some sensitive cases, thereby impeding collaboration among organizations seeking to collectively develop robust GAD models. Although federated learning offers a promising solution, the prevalent non-IID problems and high communication costs present significant challenges, particularly pronounced in collaborations with graph data distributed among different participants. To tackle these challenges, we propose an effective federated graph anomaly detection framework (FGAD). We first introduce an anomaly generator to perturb the normal graphs to be anomalous, and train a powerful anomaly detector by distinguishing generated anomalous graphs from normal ones. Then, we leverage a student model to distill knowledge from the trained anomaly detector (teacher model), which aims to maintain the personality of local models and alleviate the adverse impact of non-IID problems. Moreover, we design an effective collaborative learning mechanism that facilitates the personalization preservation of local models and significantly reduces communication costs among clients. Empirical results of the GAD tasks on non-IID graphs compared with state-of-the-art baselines demonstrate the superiority and efficiency of the proposed FGAD method.



\end{abstract}





\maketitle

\section{Introduction}
\label{sec1}
Anomaly detection~\cite{chandola2009anomaly, pang2021deep} is a fundamental research problem in machine learning, and it has been extensively explored in various domains such as images~\cite{li2021cutpaste, cai2022perturbation} and time-series data~\cite{deng2021graph, blazquez2021review, liu2022time}. In the real world, graph-structured data is commonly available due to its exceptional ability to represent complicated relationship information among entities~\cite{zhuang2018dual}. This is particularly evident in domains like social networks and medical applications. Consequently, graph anomaly detection (GAD)~\cite{ma2021comprehensive, ding2021few}, which aims to identify graphs that exhibit significant deviations from other normal graphs, has raised broad attention in recent years. With the advancement of graph neural networks (GNNs)~\cite{kipf2017semi, xu2019powerful}, GAD has made remarkable strides and demonstrated promising performance in detecting anomalies across many real-world scenarios with natural graph-structured data, e.g., social networks, molecules, and bioinformatics.

\begin{figure}[htbp]
    \centering
    \includegraphics[width=1\linewidth]{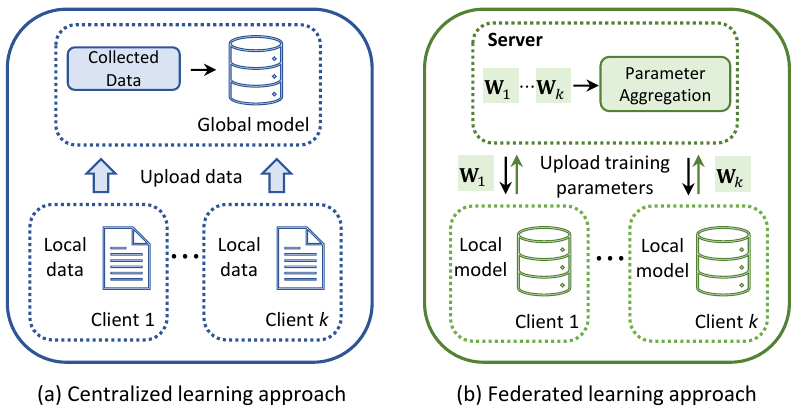}
    \caption{Overview of the centralized learning and federated learning frameworks.}
    \label{Explain}
\end{figure}

In realistic collaborative efforts among different companies and organizations, they attempt to share knowledge with each other in order to more accurately detect anomalies. However, existing GAD approaches~\cite{ma2022deep, zhang2022dual,tang2022rethinking, gao2023alleviating} typically involve a centralized model that requires all participants to provide their own data for training a global model, as shown in Figure~\ref{Explain}(a). Although this centralized training simplifies coordination, it introduces a critical privacy leakage risk. Graph data may encompass some sensitive information that the participant is not willing to share, e.g., the private relationship in social networks, which then hinders their collaborations. Consequently, an urgent imperative emerges to investigate approaches
that facilitate collaboration between GAD models distributed to different participants while protecting their privacy.

\begin{figure*}[t]
    \centering
    \includegraphics[width=0.92\linewidth]{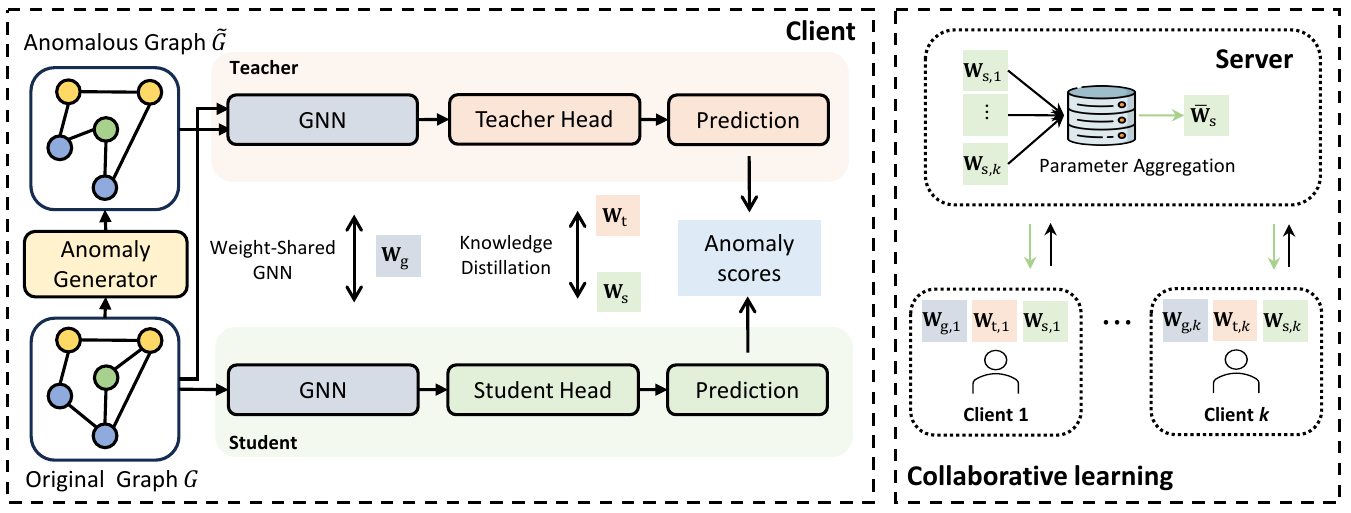}
    \caption{Overview of the FGAD framework. Note that the teacher model utilizes both normal and generated anomalous graphs for training an anomaly detector, while the student model only inputs normal graphs for the distillation of normal patterns.}
    \label{Network}
\end{figure*}
As the emerging technique in machine learning, federated learning (FL), as shown in Figure~\ref{Explain}(b), enables the collaboration between different participants with the consideration of privacy-preserving. The clients in FL only need to share their network parameters with the server rather than their local data, which prevents the leakage of sensitive information in participants. Classical FL methods, such as FedAvg~\cite{mcmahan2017communication} and FedProx~\cite{li2020fedprox}, have become the paradigm of collaborative learning across various domains~\cite{li2021meta, wu2021hierarchical}. To facilitate the collaborative training of GNN models for graph data across clients, federated graph learning (FGL)~\cite{zhang2021federated, fu2022federated, liu2022federated} has also been widely studied in recent years. FGL methods~\cite{xie2021federated, tan2023federated} integrate GNNs with FL methods to collaboratively learn representations for complicated graph data distributed in various clients, and have demonstrated superiority in graph classification tasks. Therefore, an intuitive approach to address the above issue is to integrate the existing advancements in FL and FGL with general anomaly detection techniques, e.g., deep one-class classification (DeepSVDD)~\cite{ruff2018deep}.

However, this solution may encounter the following challenges:
\begin{enumerate}[leftmargin=*]
\item The graph data distributed in various clients often exhibits significant heterogeneity and non-IID property~\cite{kairouz2021advances,xie2021federated}, e.g., containing different graph structures or feature dimensions. These factors place a higher demand on maintaining the validity of the local models for their own data, e.g., personalization.
\item It is difficult to learn a universal hypersphere as the decision boundary for highly heterogeneous graph data under the federated learning setting. Besides, such non-IID graphs across clients hardly conform to the assumption in DeepSVDD that their latent distribution could follow a universal hypersphere.
\item Existing collaborative learning mechanisms, e.g., FedAvg~\cite{mcmahan2017communication}, require transmitting all network parameters of each client in a single communication round, which brings substantial communication costs in applications.
\end{enumerate}
Those challenges naturally lead to a research question: \textit{\textbf{Can we design an FL-based GAD framework that facilitates more effective collaboration and achieves more accurate detection?}}

In this paper, we propose an effective federated graph anomaly detection (FGAD) framework, as shown in Figure~\ref{Network}, to answer this research question. To improve the anomaly detection capability in the local model, we introduce an anomaly generator that perturbs normal graphs to be anomalous, and train a classifier to identify anomalies from normal graphs. The generated anomalous graphs are encouraged to be diverse and resemble normal ones through iterations, so that more robust decision boundaries can be learned in a self-boosted manner. To alleviate the adverse impact of non-IID problems, we propose to preserve the personalization of each client by leveraging knowledge distillation. Specifically, we introduce a student model to distill the knowledge from the trained classifier (teacher model). The student model only takes the normal graphs as the input, with the aim of aligning its predicted distributions with that of the teacher model. Moreover, we further design an effective collaborative learning mechanism. We let the student and teacher models share the same backbone network to streamline the capacity of local models. Besides, we engage only the parameters of the student head rather than the entire model in collaborative learning, which allows the teacher model to preserve the personalization of a client. In this way, we not only alleviate the adverse impact of non-IID property, but also reduce the communication costs between clients and server during collaborative learning. The contributions of this paper are summarized as follows:
\begin{itemize}[leftmargin=*]
    \item We investigate the challenging anomaly detection issue on non-IID graphs distributed across various clients, and propose an effective federated graph anomaly detection (FGAD) framework.
    \item We introduce a self-boosted distillation module, which not only promotes the detecting capability by identifying self-generated anomalies, but also maintains the personalization of local models from knowledge distillation to alleviate non-IID problems.
    \item We propose an effective collaborative learning mechanism that streamlines the capacity of local models and reduces communication costs with the server.
    \item We establish a comprehensive set of baselines for federated graph anomaly detection. Extensive experiments also validate the effectiveness of the proposed FGAD method.
\end{itemize}

\section{Related Works}
\label{sec2}

\subsection{Graph Anomaly Detection}
\label{sec2.1}
Graph anomaly detection (GAD)~\cite{ma2021comprehensive} refers to detecting abnormal graphs that significantly differ from other normal ones, which have received growing attention in recent years owing to the ubiquitous prevalence of graph-structured data in real-world scenarios, such as social networks~\cite{liu2021anomaly}. There are many works that advance the research on GAD. For instance, Zhao et al.~\cite{zhao2021using} investigated graph-level anomaly detection issues by integrating graph isomorphism network (GIN)~\cite{xu2019powerful} with deep one-class classification (DeepSVDD)~\cite{ruff2018deep}. Qiu et al.~\cite{qiu22raising} leveraged neural transformation learning to develop a more robust GAD model to overcome the performance flip issue. Ma et al.~\cite{ma2022deep} utilized knowledge distillation to capture more comprehensive normal patterns from the global and local views for detecting graph anomalies.

Although these GAD methods have achieved remarkable success, they primarily rely on centralized training paradigms. Nevertheless, in real-world collaborative scenarios, the graph data is often distributed across various clients, which necessitates the transmission of local graph data to a central server during practical collaborations. Unfortunately, this process can potentially expose sensitive information and pose severe privacy risks. Additionally, the inherent non-IID property in the graph data distributed across diverse clients presents yet another formidable challenge. Consequently, the pursuit of effective solutions to address these challenges remains an open research problem.

\subsection{Federated Graph Learning}
\label{sec2.2}
Federated learning (FL) approaches~\cite{li2020federated, kairouz2021advances, yang2021characterizing}, such as FedAvg~\cite{mcmahan2017communication}, FedProx~\cite{li2020fedprox}, provide a promising solution
for collaboratively training models with data distributed in different clients, while preserving their privacy. In FL, clients only share their network parameters rather than data with the central server, which mitigates the privacy leakage risk and enables clients to share and leverage knowledge from others. As an emerging technique, FL has not only made remarkable advancements in image~\cite{li2021model, elmas2022federated,yan2023label} and time series data~\cite{wang2022system, liu2020deep}, but also raised increasing attention to graph data~\cite{zhang2021federated, wang2022federatedscope}, where collaborative efforts are significantly more challenging due to the complex structural information and heterogeneous characteristic of graphs compared to other data types.

Federated graph learning (FGL)~\cite{zhang2021federated} aims to facilitate the collaboration of GNNs distributed in multiple remote clients to meet the requirement of handling complicated non-IID graph data that widely exist in many real-world scenarios, e.g., social networks, medical, and biological data. For example, Xie et al.~\cite{xie2021federated} studied the federated learning issue on non-IID graphs by integrating the clustered federated learning with graph isomorphism network (GIN), which achieves effective collaborations for distributed GINs. Tan et al.~\cite{tan2023federated} designed a structural knowledge-sharing mechanism to facilitate the federated graph learning process. However, existing FGL methods have primarily been validated for graph classification tasks, and their effectiveness in addressing the intricate unsupervised task of graph anomaly detection remains an area of ongoing exploration. While it is possible to extend these FL/FGL~\cite{mcmahan2017communication, li2020fedprox, xie2021federated, tan2023federated} methods to address GAD tasks by integrating them with classical solutions like DeepSVDD~\cite{ma2021comprehensive,ruff2018deep}, it is imperative to acknowledge some significant challenges, e.g., the adverse impact of the non-IID problem across different clients and the communication costs of transmitting complex GNN model parameters during collaborative learning.

\section{Methodology}
\label{sec3}
\subsection{Preliminary and Problem Formulation}
\label{sec3.1}
\paragraph{\textbf{Notation:}} Let $D = \{G_{1}, \dots, G_{N} \}$ denotes a graph dataset which consists of $N$ graphs, and each graph $G_{i} = \{V_{i}, E_{i}\}$ in the graph set comprises a node set $V_{i}$ and edge set $E_{i}$. Typically, assume the number of nodes in a graph $G_{i}$ is $n_{i} = |V_{i}|$, an adjacency matrix $\mathbf{A}_{i} \in \{0, 1\}^{n_{i}\times n_{i}}$ is used to represent the topology of graph $G_{i}$. Besides, let $\mathbf{x}_{v} \in \mathbb{R}^{d}$ denotes the attribute vector for node $v\in V_{i}$, $\mathbf{X}_{i} \in \mathbb{R}^{n_{i}\times d}$ is used to represent the attribute matrix of graph $G_{i}$.

\paragraph{\textbf{Graph Neural Networks:}} Graph neural networks (GNNs), which iteratively learn representations with neighborhood aggregation and message propagation, is a widely used paradigm of learning representation for graph-structured data in many downstream tasks. In this paper, we leverage the graph isomorphism network (GIN)~\cite{xu2019powerful}, a widely used GNN backbone, to learn graph representation for anomaly detection tasks. Generally, in each layer of a GIN, the node representation is updated by aggregating its neighborhood information. For instance, in the $k$-th layer of GIN, the learned aggregated features $\mathbf{a}_{v}^{(k)}$ for node $v$ can be formulated as:
\begin{eqnarray}
\mathbf{a}_{v}^{(k)}= \mathrm{AGGREGATE}(\{\mathbf{h}^{(k-1)}(u),u \in\tilde{\mathcal{N}}(v)\}),
\label{Aggregation}
\end{eqnarray}
where $\mathrm{AGGREGATE}(\cdot)$ indicates the aggregation function, and $\tilde{\mathcal{N}}(v)$ represents the neighbor node set of node $v$. Then, the node feature $\mathbf{h}_{v}^{(k)}$ of node $v$ in the $k$-th layer is obtained by combing the node feature learned in the $(k-1)$-th layer with the aggregated feature, i.e.:
\begin{eqnarray}
\mathbf{h}_{v}^{(k)}= \sigma(\mathrm{COMBINE}(\mathbf{h}_{v}^{(k-1)},\mathbf{a}_{v}^{(k)})),
\label{Combination}
\end{eqnarray}
where $\sigma(\cdot)$ denotes the activation function, e.g., LeakyReLU. Particularly, the initial feature $\mathbf{h}_{v}^{(0)}$ for node $v$ is set as $\mathbf{h}_{v}^{(0)} = \mathbf{x}_{v}$. Consequently, we can obtain the representation for a graph $G$ based on the learned features of all nodes within $G$ as follows:
\begin{eqnarray}
\mathbf{h}_{G}= \mathcal{R}(\mathrm{CONCAT}(\mathbf{h}_{v}^{(k)}, k \in \{1, \dots, K\}), v\in G),
\label{Readout}
\end{eqnarray}
where $K$ is the number of GIN layers, and $\mathrm{CONCAT}(\cdot)$ denotes the concatenate operation that stacks the graph representation learned across all $K$ layers. $\mathcal{R}(\cdot)$ denotes the readout function that obtains the graph-level representation by aggregating the node features within a graph, and we choose sum-readout in this paper. Note that for convenience, we use $\mathrm{GIN}(\cdot)$ to simply represent a GIN model containing the above three operations, in the following sections.

\paragraph{\textbf{Problem Formulation:}} The objective of the GAD under the FL setup is to facilitate collaboration among clients, which allows each participant to enhance their GAD models by leveraging knowledge from others without exposing private data. Given $C$ clients, the collective graph dataset is denoted as $D = \{D_{1}, \dots, D_{C}\}$, where each client possesses its own graph set $D_{c}$. A prevalent paradigm in GAD~\cite{ma2021comprehensive} is that all graphs within the client, i.e., $\forall G_{i} \in D_{c}$, are deemed as ``normal''. The model is trained to capture this normality so that the trained model can distinguish an ``anomalous" graph $\tilde{G}$ deviates significantly from the distribution of $D_{c}$ by some pre-defined assumptions, e.g., the hypersphere decision boundary in DeepSVDD~\cite{ruff2018deep}. Conversely, in this paper, we attempt to develop a classifier-based anomaly detector, which can adaptively learn decision boundaries rather than relying on the strong assumption of the shape of the latent distribution. This can be regarded as solving the following problem:
\begin{eqnarray}
\underset{\mathbf{w}^{(1)}, \dots, \mathbf{w}^{(C)}}{\mathrm{minimize}}\ \  \frac{1}{C} \sum_{c=1}^{C} \frac{|D_{c}|}{|D|}( \ell_{c}(y, f_{\mathbf{w}^{(c)}}(G)) + \ell_{c}(\tilde{y}, f_{\mathbf{w}^{(c)}}(\tilde{G})) ) ,
\label{Overall_objective}
\end{eqnarray}
where $|D|$ and $|D_{c}|$ denote the total number of graphs and that of in $c$-th client. $\{G; y\}$ represents the normal graph labeled with $y=1$, and $\{\tilde{G}, \tilde{y}\}$ represents the anomalous graph labeled with $y=0$. $\ell_{c}(\cdot)$ denotes the local loss function of $c$-th client, e.g., binary cross-entropy loss. $f_{\mathbf{w}^{(c)}}(\cdot)$ is the GIN-based neural network of $c$-th client, which is parameterized by $\mathbf{w}^{(c)}$. However, tackling this problem presents the following challenges:

\begin{itemize}[leftmargin=*]
\item[1)] GAD is generally an unsupervised task that only normal graph $\{G; y\}$ is accessible. Thus, how to produce high-quality anomalous graph $\{\tilde{G}; \tilde{y}\}$ for training each local anomaly detector?
\item[2)] In the context of FL-based GAD, how to alleviate the adverse impact of the non-IID property that is prevalent in the graph data across clients?
\item[3)] Transmitting all network parameters following conventional FL methods may limit the scalability given the complexity of GIN. Therefore, how to reduce communication costs in collaborative learning while maintaining the validity of local models?
\end{itemize}

\subsection{Self-boosted Graph Knowledge Distillation}
\label{sec3.2}
The first challenge raises the demand to produce anomalous graphs without using any supervised information. To this end, we propose a graph anomaly generator denoted as $\mathcal{G}_{\mathbf{w}_{a}}(\cdot)$ to generate anomalous graphs by perturbing the graph structure of normal graph $G$. For each client, we aim to generate an anomalous graph set $\tilde{D}_{c} = \{\mathbf{X}_{c}, \tilde{\mathbf{A}}_{c}\}$ in an unsupervised manner by feeding with normal graph set $D_{c}$. To ensure diversity in the generated anomalous graphs, we leverage variational graph auto-encoder (VGAE)~\cite{kipf2016variational} to build the anomaly generator. Specifically, we first learn a latent Gaussian distribution $\mathcal{N}(\bm{\mu}_{c}, \bm{\sigma}_{c}^{2})$, which can be determined as follows:
\begin{eqnarray}
\bm{\mu}_{c} = \mathrm{GIN}_{\bm{\mu}} (\mathbf{X}_{c}, \mathbf{A}_{c}), \bm{\sigma}_{c} = \mathrm{GIN}_{\bm{\sigma}} (\mathbf{X}_{c}, \mathbf{A}_{c}),
\label{mu_sigma}
\end{eqnarray}
where $\mathrm{GIN}_{\bm{\mu}}(\cdot)$ and $\mathrm{GIN}_{\bm{\sigma}}(\cdot)$ denotes two distinct GINs in anomaly generator, and $\bm{\mu}_{c}$ and $\bm{\sigma}_{c}$ explicitly parameterize the following inference model:
\begin{eqnarray}
q(\tilde{\mathbf{Z}}_{c}|\mathbf{X}_{c},\mathbf{A}_{c}) = \prod_{i=1}^{|D_{c}| }q(\mathbf{Z}_{c}^{(i)}|\mathbf{X}_{c},\mathbf{A}_{c}),
\label{inference}
\end{eqnarray}
where $q(\tilde{\mathbf{Z}}_{c}^{(i)}|\mathbf{X}_{c},\mathbf{A}_{c})=\mathcal{N}(\tilde{\mathbf{Z}}_{c}^{(i)}|\bm{\mu}_{c}^{(i)}, \textrm{diag}(\bm{\sigma}_{c}^{(i)}) )$, and it allows us to sample from a wide range in the latent space thereby facilitating the diverse anomalous graph generation. Here, we employ the reparametrization trick~\cite{kingma2013auto} to address the obstacle of gradient propagation in the sample operation. Consequently, the generated adjacency matrix can be calculated by:
\begin{eqnarray}
\tilde{\mathbf{A}}_{c} = \mathcal{T}(\tilde{\mathbf{Z}}_{c}^{\top}\tilde{\mathbf{Z}}_{c}), \ \ \tilde{\mathbf{Z}}_{c} = \bm{\mu}_{c} + \epsilon \bm{\sigma}_{c}, \ \epsilon \sim \mathcal{N}(\bm{0},\bm{1}),
\label{re-para}
\end{eqnarray}
where $\mathcal{T}:\mathbb{R} \to [0, 1]$ represents the element-wise transformation operations such as $\textrm{Sigmoid}(\cdot)$, and $\epsilon$ represents a random Gaussian noise that follows the standard normal distribution $\mathcal{N}(\bm{0}, \bm{1})$.

Intuitively, allowing the generated graphs to closely resemble normal graphs while remaining as anomalies is beneficial in training a robust and powerful anomaly detector, as it forces the model to distinguish those subtle deviations from the normal patterns. Therefore, we propose to optimize the anomaly generator by minimizing the following objective:
\begin{eqnarray}
\ell_{\text{g}}^{c}(\mathbf{A}_{c}, \tilde{\mathbf{A}}_{c}) = -\sum_{i,j} (\mathbf{A}_{c}^{ij} \log(\tilde{\mathbf{A}}_{c}^{ij}) + (1 - \mathbf{A}_{c}^{ij}) \log(1 - \tilde{\mathbf{A}}_{c}^{ij})),
\label{loss_aug}
\end{eqnarray}
where $\ell_{\mathrm{g}}^{c}$ denotes the binary-cross entropy loss function. Subsequently, we can train an anomaly detector with the normal and generated anomalous graph sets for the local client as follows:
\begin{align}
\ell_{\mathrm{ad}}^{c} = l_{\mathrm{ce}}(y_{c},\mathrm{Proj}&(f_{\mathbf{w}_{g}}(\mathbf{X}_{c}, \mathbf{A}_{c})))+l_{\mathrm{ce}}(\tilde{y}_{c},\mathrm{Proj}(f_{\mathbf{w}_{g}}(\mathbf{X}_{c}, \tilde{\mathbf{A}}_{c}))), \label{loss_anomaly_detector}
\end{align}
where $\tilde{\mathbf{A}}_{c} = \mathcal{G}_{\mathbf{w}_{a}}(\mathbf{X}_{c},\mathbf{A}_{c})$, $l_{\mathrm{ce}}(\cdot)$ is the cross-entropy loss, and $f_{\mathbf{w}_{g}}(\cdot)$ denotes the GIN backbone that learns graph representation by feeding with graph data. $\mathrm{Proj}(\cdot)$ is the MLP-based projection head that maps the graph representation learned from $f_{\mathbf{w}_{g}}(\cdot)$ into the predicted logits. Note that we simply set the label of the normal graph $y_{c} = 1$, and the generated anomalous graph as $\tilde{y}_{c} = 0$.

Hence, we can train an anomaly detector in an unsupervised manner by minimizing the following objective function:
\begin{eqnarray}
\ell_{\mathrm{pt}} = \frac{1}{C} \sum_{c=1}^{C} \frac{|D_{c}|}{|D|}(\ell_{\mathrm{ad}}^{c} + \ell_{\mathrm{g}}^{c}),
\label{pre-train}
\end{eqnarray}
where $\ell_{\mathrm{g}}^{c}$ attempts to generate anomalous graphs that closely resemble normal ones, while
$\ell_{\mathrm{ad}}^{c}$ aims to identify those generated anomalous graphs. Therefore, we produce diverse anomalous graphs for learning a powerful anomaly detector in such a self-boosted style, and the two objectives mutually improve each other during training.

However, in the context of federated learning, the graph data across different clients is often heterogeneous and exhibits non-IID property. Such characteristics can potentially affect the anomaly detection performance of local models, i.e., the second challenge. To alleviate the adverse impact of the non-IID problem, we propose a graph knowledge distillation framework, which is designed to preserve the personalization of the local model during collaborative learning. Specifically, we regard the previously pre-trained anomaly detector as the ``teacher'' model, and introduce a ``student'' model that aims to distill the knowledge from the teacher model and achieve collaboration between clients.

The network architecture of the student model is similar to the teacher model, which consists of a GIN backbone and a projection head. Since the purpose of the student model is to mimic the predictions of the teacher model for normal data, only normal graphs are considered in the knowledge distillation. The predicted logits of the teacher and student models are computed as follows:
\begin{eqnarray}
\mathbf{Q}_{c, \mathrm{t}}=\mathrm{Proj}_{\mathrm{t}|\mathbf{w}_{t}}(f_{\mathbf{w}_{g}}(\mathbf{X}_{c}, \mathbf{A}_{c})), \ \ \mathbf{Q}_{c, \mathrm{s}}=\mathrm{Proj}_{\mathrm{s}|\mathbf{w}_{s}}(f_{\mathbf{w}_{g'}}(\mathbf{X}_{c}, \mathbf{A}_{c})),
\label{Logits}
\end{eqnarray}
where $f_{\mathbf{w}_{g}}(\cdot)$, $\mathrm{Proj}_{\mathrm{t}|\mathbf{w}_{t}}(\cdot)$ and $f_{\mathbf{w}_{g'}}(\cdot)$, $\mathrm{Proj}_{\mathrm{s}|\mathbf{w}_{s}}(\cdot)$ are the backbone networks and projection heads of teacher and student models respectively. Note that $\mathrm{Proj}_{\mathrm{t}|\mathbf{w}_{t}}(\cdot)$ is actually the same as the projection head $\mathrm{Proj}(\cdot)$ in Eq.~\eqref{loss_anomaly_detector}. Subsequently, the student model distills the knowledge from the teacher model by matching its predicted logits with those of the teacher model, described as follows:
\begin{eqnarray}
\ell_{\text{kd}}^{c} = \frac{1}{|D_{c}|}\sum_{i\in D_{c}}KL(\mathrm{softmax}(\mathbf{Q}_{c, \mathrm{t}}^{(i)}/\tau), \mathrm{softmax}(\mathbf{Q}_{c, \mathrm{s}}^{(i)}/\tau)),
\label{Loss_distillation}
\end{eqnarray}
where $KL(\cdot,\cdot)$ denotes the Kullback-Leibler divergence, which is applied to measure the discrepancy between the distribution of the predicted logits from teacher and student models. $\mathrm{softmax}(\cdot)$ is the softmax function, i.e., $\mathrm{softmax}(q_{i}/\tau)= \frac{\exp(q_{i}/\tau)}{\sum_{j}\exp(q_{j}/\tau)}$, and $\tau$ is the temperature factor that controls the smoothness of the distillation.

\subsection{Parameter-efficient Collaborative Learning}
\label{sec3.3}
Based on the design of the self-boosted graph knowledge distillation module, the objective function of all clients is defined as follows:
\begin{eqnarray}
\mathcal{L}_{\mathrm{total}} = \frac{1}{C} \sum_{c=1}^{C} \frac{|D_{c}|}{|D|}(\ell_{\mathrm{ad}}^{c} + \lambda \ell_{\mathrm{g}}^{c} + \gamma \ell_{\text{kd}}^{c}),
\label{Total_loss}
\end{eqnarray}
where $\lambda$ and $\gamma$ are the two trade-off parameters. In federated learning, let $\mathbf{W}^{(c)} = \{\mathbf{w}_{a}^{(c)}, \mathbf{w}_{g}^{(c)}, \mathbf{w}_{g'}^{(c)}, \mathbf{w}_{t}^{(c)}, \mathbf{w}_{s}^{(c)}\}$ denotes the parameter set of the $c$-th client, the conventional solution achieves collaboration by uploading the network parameters to the server and then distribute the aggregated network parameters to each client. However, this solution presents several problems. First, the high parameter complexity of a GIN-based backbone can limit the scalability of the model during the parameter aggregation process. Second, the transmission of all network parameters may introduce non-IID problems, and affect the performance of local models trained on different graph data across clients.

To address these issues, we propose an effective collaborative learning mechanism in this paper, which is described in Figure~\ref{Network}. Specifically, We let the teacher and student models share the same GIN backbone for learning graph representation, i.e.,
\begin{eqnarray}
\mathbf{Z}_{c} = f_{\mathbf{w}_{g}}(\mathbf{X}_{c}, \mathbf{A}_{c}) = f_{\mathbf{w}_{g'}}(\mathbf{X}_{c}, \mathbf{A}_{c}),
\label{Shared_feature}
\end{eqnarray}
where $\mathbf{Z}_{c}$ denotes the learned graph representation that is shared as the input to the projection heads of teacher and student. This operation not only reduces the complexity of the local model, but also simplifies the knowledge distillation of the student model. Then we only upload the parameter set $\mathbf{w}_{s}^{(c)}$ of the student head for collaboration instead of uploading all the network parameters, i.e., the parameter aggregation in the server is formalized as follows:
\begin{eqnarray}
\bar{\mathbf{w}}_{s} = \sum_{c=1}^{C} \frac{|D_{c}|}{|D|} \mathbf{w}_{s}^{(c)},
\label{Collabrative_learning}
\end{eqnarray}
where $\bar{\mathbf{w}}_{s}$ denotes the aggregated parameters in the server. The proposed collaborative learning mechanism not only streamlines the capacity of local models, but also significantly reduces the communication costs, which addresses the third challenge. To facilitate the understanding of the proposed FGAD method, we summarize its detailed training process in Algorithm~\ref{algorithm}. The collaboration between clients via the student model is performed in the following two steps:
\begin{itemize}[leftmargin=*]
\item Each client performs graph knowledge distillation independently, updating its network parameters, and uploads the network parameters of the student head to the server.
\item The server then aggregates the network parameters following Eq.~\ref{Collabrative_learning}, and distributes the aggregated network parameters to each client.
\end{itemize}

\begin{algorithm}[h]
	\caption{Training process of the proposed FGAD}
	\label{algorithm}
	\begin{algorithmic}[1]
		\REQUIRE {Graph set $D = \{D_{c}\}_{c=1}^{C}$, number of clients $C$, number of GNN layers $K$, learning rate $\alpha$, total epochs $\mathcal{T}$.}
		\ENSURE {The overall graph anomaly detection performance.}
		\STATE {Initialize the parameter sets $\{\mathbf{W}^{(c)}\}_{c=1}^{C}$ for each local model;}
            \STATE {Pretrain the local model in each client with Eq.~\eqref{pre-train};}
            \WHILE{not converge}
		\FOR{$t=1, 2, \dots, \mathcal{T}$}
            \FOR{$c =1, \dots, C$}
            \STATE {Generate anomalous graph set $\tilde{\mathbf{D}}$ with Eqs.~\eqref{mu_sigma}, \eqref{inference}, \eqref{re-para};}
            \STATE {Compute loss items $\ell_{\mathrm{ad}}^{c}$, $\ell_{\mathrm{g}}^{c}$, $\ell_{\mathrm{kd}}^{c}$ with Eq.~\eqref{loss_aug}, \eqref{loss_anomaly_detector}, \eqref{Loss_distillation};}
		\ENDFOR
            \STATE {Back-propagation and update each local model via minimizing Eq.~\eqref{Total_loss};}
            \ENDFOR
            \STATE {Upload the parameter sets $\{\mathbf{w}_{s}^{(c)}\}_{c=1}^{C}$ of student model in each client to the server; }
            \STATE {Compute aggregated network parameters $\bar{\mathbf{w}}_{s}$ with collaborative learning following Eq.~\eqref{Collabrative_learning}; }
            \STATE {Distribute parameter set $\bar{\mathbf{w}}_{s}$ to the local model of each client;}
            \ENDWHILE
		\STATE {Evaluate the anomaly detection performance in each client and aggregate their results; }
		\RETURN {The overall graph anomaly detection performance.}
	\end{algorithmic}
\end{algorithm}

\section{Experiment}
\label{sec4}

\subsection{Experimental Setup}
\label{sec4.1}

\paragraph{\textbf{Datasets}}
We evaluate the performance of FL-based graph anomaly detection on non-IID graphs through two distinct experimental setups: (1) single-dataset and (2) multi-dataset scenarios.
\begin{itemize}[leftmargin=*]
\item \textbf{Single-dataset:} we distribute a single dataset across multiple clients, each of which possesses a unique subset of the dataset. This setup allows us to assess the effectiveness when clients collaborate on a shared dataset. We employ three social network datasets including IMDB-BINARY, COLLAB, and IMDB-MULTI to conduct this experiment.
\item \textbf{Multi-dataset:} we broaden our evaluation by considering various datasets distributed in multiple clients and each of them holds a specific dataset. We consider not only social network data (SOCIALNET) but also expand to include molecular (MOLECULES), biochemical (BIOCHEM), and mix data types (MIX). This allows us to thoroughly assess FL-based graph anomaly detection across a spectrum of data types and collaboration scenarios.
\end{itemize}
The information and construction details of each dataset are illustrated in Appendix~\ref{Appendix_data}.

\begin{table*}[t]
\centering
\setlength{\tabcolsep}{2.85mm}{
\renewcommand{\arraystretch}{1}
\caption{Anomaly detection performance (mean(\%) $\pm$ std(\%)) under the single-dataset setting. Note that the best performance is marked in Bold, and the last column shows the number of transmitted parameters in collaborative learning.}
\label{result}
\begin{tabular}{l||c|c|c|c|c|c|c}
\toprule
\multirow{2}{*}{Methods} & \multicolumn{2}{c|}{IMDB-BINARY} & \multicolumn{2}{c|}{COLLAB} &\multicolumn{2}{c|}{IMDB-MULTI}  &\multirow{2}{*}{\# Parameters} \\ \cmidrule{2-7}
& AUC & AUPRC & AUC & AUPRC & AUC & AUPRC &\\ \midrule
Self-train & 41.58$\pm$1.34& 47.43$\pm$1.39&46.96$\pm$1.80 &30.87$\pm$0.62 & 52.39$\pm$1.31&32.74$\pm$0.60   & N/A\\
\midrule
FedAvg~\cite{mcmahan2017communication}      & 40.96$\pm$3.44 & 48.24$\pm$2.41 & 49.60$\pm$0.45 & 30.69$\pm$0.50 & 49.11$\pm$1.46& 36.13$\pm$1.54  & 5,370,880 \\
FedProx~\cite{li2020fedprox}     & 39.62$\pm$2.36 & 46.74$\pm$1.24 & 49.56$\pm$0.50 & 31.40$\pm$0.50&52.16$\pm$1.75 & 36.13$\pm$1.54 & 5,370,880 \\
\midrule
GCFL~\cite{xie2021federated}        & 56.98$\pm$5.56 & 59.68$\pm$3.37 & 48.93$\pm$1.02 & 30.84$\pm$0.36&49.44$\pm$2.95 & 34.87$\pm$0.68   & 10,741,760 \\
FedStar~\cite{tan2023federated}    & 54.76$\pm$1.28& 56.49$\pm$0.86& 51.89$\pm$0.33& 36.89$\pm$0.43& 58.28$\pm$0.53& 39.97$\pm$1.22 &  416,000\\
\midrule
FGAD        & \bf64.97$\pm$0.52 & \bf66.60$\pm$1.12 & \bf55.08$\pm$1.85 & \bf66.67$\pm$0.00&\bf60.51$\pm$1.18& \bf66.82$\pm$0.14 & 21,130 \\
\bottomrule
\end{tabular}}
\end{table*}

\begin{table*}[t]
\centering
\caption{Anomaly detection performance (mean(\%) $\pm$ std(\%)) under the multi-dataset setting. Note that the best performance is marked in Bold.}
\label{result2}
\setlength{\tabcolsep}{2.1mm}{
\renewcommand{\arraystretch}{1}
\begin{tabular}{l||c|c|c|c|c|c|c|c}
\toprule
\multirow{2}{*}{Methods} & \multicolumn{2}{c|}{MOLECULES} &\multicolumn{2}{c|}{BIOCHEM} &\multicolumn{2}{c|}{SOCIALNET} &\multicolumn{2}{c}{MIX}\\ \cmidrule{2-9}
& AUC & AUPRC & AUC & AUPRC & AUC & AUPRC & AUC & AUPRC \\ \midrule
Self-train
&61.26$\pm$2.91&61.31$\pm$1.91&54.54$\pm$0.99&52.29$\pm$0.40& 50.31$\pm$1.55
& 39.96$\pm$1.58&51.94$\pm$0.42 &47.65$\pm$0.64\\
\midrule
FedAvg~\cite{mcmahan2017communication}      & 54.41$\pm$3.21
& 55.55$\pm$3.23& 40.88$\pm$1.36&51.63$\pm$1.13  & 48.21$\pm$1.02& 38.29$\pm$1.29& 47.96$\pm$0.61&44.89$\pm$0.68\\
FedProx~\cite{li2020fedprox}     & 57.93$\pm$2.14&58.72$\pm$2.25 & 46.04$\pm$0.49&51.57$\pm$0.80 & 47.26$\pm$0.10&37.23$\pm$0.92 &46.79$\pm$0.63 &44.19$\pm$0.29\\
\midrule
GCFL~\cite{xie2021federated}         &45.67$\pm$1.33& 51.96$\pm$0.79& 41.49$\pm$0.30&52.23$\pm$0.65 & 47.59$\pm$0.95& 37.53$\pm$0.93 & 49.58$\pm$0.50&45.37$\pm$0.69\\
FedStar~\cite{tan2023federated}    & 56.15$\pm$0.92
& 59.73$\pm$1.21& 47.80$\pm$0.48&56.48$\pm$0.19 & 53.79$\pm$2.03& 36.40$\pm$1.11& 50.53$\pm$1.11&45.83$\pm$0.41\\
\midrule
FGAD        &\bf62.15$\pm$0.69 &\bf79.19$\pm$0.49 & \bf58.09$\pm$0.85& \bf59.04$\pm$0.54& \bf54.86$\pm$0.29& \bf56.88$\pm$0.98&\bf58.14$\pm$0.36 &\bf52.03$\pm$0.63\\
\bottomrule
\end{tabular}}
\end{table*}

\paragraph{\textbf{Baseline Methods}} We compare the proposed FGAD method with several state-of-the-art baseline methods. We include two federated learning methods: FedAvg~\cite{mcmahan2017communication} and FedProx~\cite{li2020fedprox}, as well as two federated graph learning methods: GCFL~\cite{xie2021federated} and FedStar~\cite{tan2023federated}. Note that in order to adapt these baseline methods to the graph anomaly detection task, we integrate them with DeepSVDD~\cite{ruff2018deep} to construct an end-to-end graph anomaly detection model. Besides, we regard the self-training strategy without the FL setting as one of the baselines. To ensure a fair comparison with FGAD, we employ the same GIN  network structure as FGAD in all baseline methods.

\paragraph{\textbf{Implementation Details}} We use GIN~\cite{xu2019powerful} as the graph representation learning backbone for FGAD and all baselines. The number of GIN layer $K$ is set to 3, and the dimensions of the hidden layer of GIN and projection head of student and teacher models are all set to 64. We use Adam~\cite{kingma2013auto} as the optimizer and fixed the learning rate $\alpha =0.001$. For all datasets, we first pretrain the anomaly generator and teacher model for 10 epochs, and then jointly train with knowledge distillation and collaborative learning for 200 epochs. For more training details, please refer to Appendix~\ref{Appendix_experimental_setting}.

\paragraph{\textbf{Evaluation Metrics:}} We use Area Under the Curve (AUC) and Area Under the Precision-Recall Curve (AUPRC) as the evaluation metrics in the experiment. Each method is executed 10 times to report their means and standard deviations.

\subsection{Experimental Results}
\label{sec4.2}
In this section, we conduct comprehensive experiments including two types of non-IID graph scenarios, i.e., the single-dataset and multi-dataset distributed in multiple clients, to validate the effectiveness of the proposed method. Table \ref{result} and Table \ref{result2} show the experimental results of FGAD and several state-of-the-art baselines, from which we can have the following observations.
\begin{itemize}[leftmargin=*]
\item \textbf{Comparison:} In the single-dataset experiment, FGAD demonstrates a remarkable advantage over all baseline methods. For instance, in the IMDB-BINARY dataset, FGAD achieves significant performance improvement, exceeding Self-train by 23.39\% in AUC and 19.17\% in AUPRC. It also significantly surpasses classical FedAvg and FedProx. Furthermore, FGAD outperforms the state-of-the-art baselines GCFL and FedStar by a substantial 7.99\% and 10.21\% in AUC, respectively. Similar trends are evident across other benchmarks, demonstrating the effectiveness of FGAD. In the multi-dataset experiment, the GAD task is more challenging as the non-IID problem in it is more severe compared to the single-dataset scenario. Nevertheless, FGAD still exhibits outstanding performance compared to other baselines. For example, on MOLECULES, FGAD outperforms the runner-up FedStar by 6\% in AUC and 19.46\% in AUPRC. Besides, it achieves more than a 10.00\% performance improvement compared to other baseline methods. More importantly, we can observe from Table~\ref{result} that FGAD significantly reduces communication costs during collaborative learning compared to other baseline methods.
\item \textbf{Discussion:} The Self-train strategy discards collaborative training and fails to leverage the knowledge from other clients to learn more robust local GAD models. FedAvg and FedProx require the transmission of all network parameters of the local models, which introduces severe non-IID problems in collaborative learning. Consequently, these three aforementioned baselines yield suboptimal performance in most cases.
Although GCFL incorporates a specific design to alleviate non-IID challenges, such as utilizing clustered FL for collaborative learning, it still necessitates the transmission of all network parameters and does not effectively address non-IID problems, as validated by the experimental results. On the other hand, FedStar achieves runner-up performance in most cases, which may primarily be attributed to the introduced structural embedding that helps to preserve the personalization of local models. Compared with the baseline methods, FGAD considers enhancing the detecting capability of local models in a self-boosted manner, and introduces an effective collaborative learning mechanism by leveraging knowledge distillation. This allows FGAD to learn more powerful local GAD models, mitigate the adverse effects of non-IID problems, and reduce communication costs among clients.
\end{itemize}

\subsection{Embedding Visualization}
\label{sec4.3}
\begin{figure}[t]
    \centering
\includegraphics[width=1\linewidth]{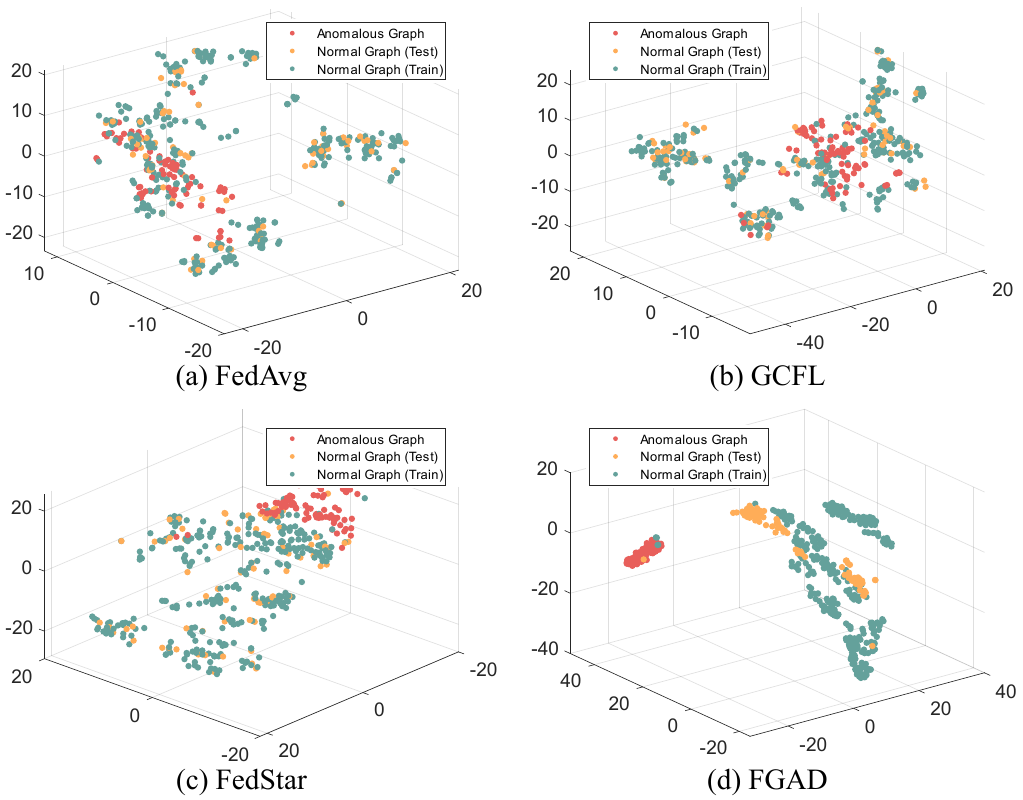}
   \caption{Embedding visualization of the proposed FGAD compared with several baselines using t-SNE. Note that the data point marked in \textcolor{Kingdom_Yellow}{yellow}, \textcolor{Hotel_Pink}{red}, and \textcolor{Aquatic_Blue}{green} correspond to the normal graph (test), anomalous graph, and normal graph (train), respectively.}
   \label{tsne_comparison}
\end{figure}
We employ t-SNE~\cite{van2008visualizing} to visualize the learned embedding for intuitive comparison. Figure \ref{tsne_comparison} shows the embedding visualization for AIDS, one of the constituents of MOLECULES. We include results from FedAvg, GCFL, and FedStar for a comprehensive analysis.
It's evident that the learned embeddings by FedAvg and GCFL exhibit poor discriminative properties, with both normal and anomalous graphs appearing entangled in the latent space.  Although the visualization result of FedStar shows some separation between normal and anomalous graphs, it remains blurred decision boundaries. Conversely, the learned embeddings of FGAD are clearly more discriminative compared to the other baseline methods. The visualization of FGAD reveals distinct boundaries between the embeddings of normal and anomalous graphs, supporting its effectiveness.

\begin{table}[t]
\centering
\caption{Ablation study results (mean(\%) $\pm$ std(\%)) of FGAD and its three variants.}
\setlength{\tabcolsep}{1mm}{
\renewcommand{\arraystretch}{1}
\begin{tabular}{l||c|c|c|c}
\toprule
\multirow{2}{*}{Methods} & \multicolumn{2}{c|}{IMDB-MULTI} &\multicolumn{2}{c}{MOLECULES} \\ \cmidrule{2-5}
& AUC & AUPRC & AUC & AUPRC\\ \midrule
FGAD\_v1&56.67$\pm$1.72&64.91$\pm$1.85&57.98$\pm$2.78&75.80$\pm$0.09 \\
FGAD\_v2  &56.69$\pm$1.22 &65.98$\pm$0.90&59.41$\pm$2.22&77.32$\pm$1.02 \\
FGAD\_v3   & 55.23$\pm$3.54&61.02$\pm$2.68&55.58$\pm$4.56&66.73$\pm$0.80 \\
\midrule
FGAD        & \bf60.51$\pm$1.18& \bf66.82$\pm$0.14&\bf62.15$\pm$0.69 &\bf79.19$\pm$0.49 \\
\bottomrule
\end{tabular}}
\label{Ablation_study}
\end{table}
\subsection{Ablation Study}
\label{sec4.4}
To validate the effectiveness of each component in the proposed FGAD method, we derive three variants from FGAD and perform a systematic evaluation. Specifically, we illustrate the construction details of the three variants as follows:
\begin{itemize}[leftmargin=*]
\item \textbf{FGAD\_v1:} This variant only considers local training in each client, and abandons the collaborative learning between clients.
\item \textbf{FGAD\_v2:} This variant drops the proposed collaborative learning mechanism, and follows the parameter aggregation mechanism of the classical FedAvg method.
\item \textbf{FGAD\_v3:} This variant drops the knowledge distillation module, i.e., removes the student model and only takes the classifier (teacher model) in collaboration.
\end{itemize}
Table~\ref{Ablation_study} shows the experimental results of FGAD and its three variants on two datasets, yielding the following observations. FGAD\_v1 demonstrates a performance decline compared to FGAD, which is primarily due to the fact that FGAD\_v1 exclusively focuses on local training, neglecting collaboration with other clients. Consequently, it fails to leverage the comprehensive knowledge of other clients. Secondly, when we substitute the proposed collaborative learning mechanism with the classical FedAvg, there is also a noticeable performance decline. This can be attributed to the potential susceptibility of parameter transmission in FedAvg to the adverse effects of non-IID problems. Third, FGAD consistently outperforms FGAD\_v3 by a significant margin. This observation reveals the crucial role of the self-boosted distillation module in maintaining the personalization of local models within each client, which effectively mitigates the non-IID problems. Overall, the ablation study results fully support the rationale and the effectiveness of each component proposed in FGAD.


\subsection{Parameter Analysis}
\label{sec4.5}
\subsubsection{\textbf{Impact of Hyper-Parameters $\lambda$ and $\gamma$}}
\label{sec4.5.1}

\begin{figure}[t]
    \centering
\includegraphics[width=1\linewidth]{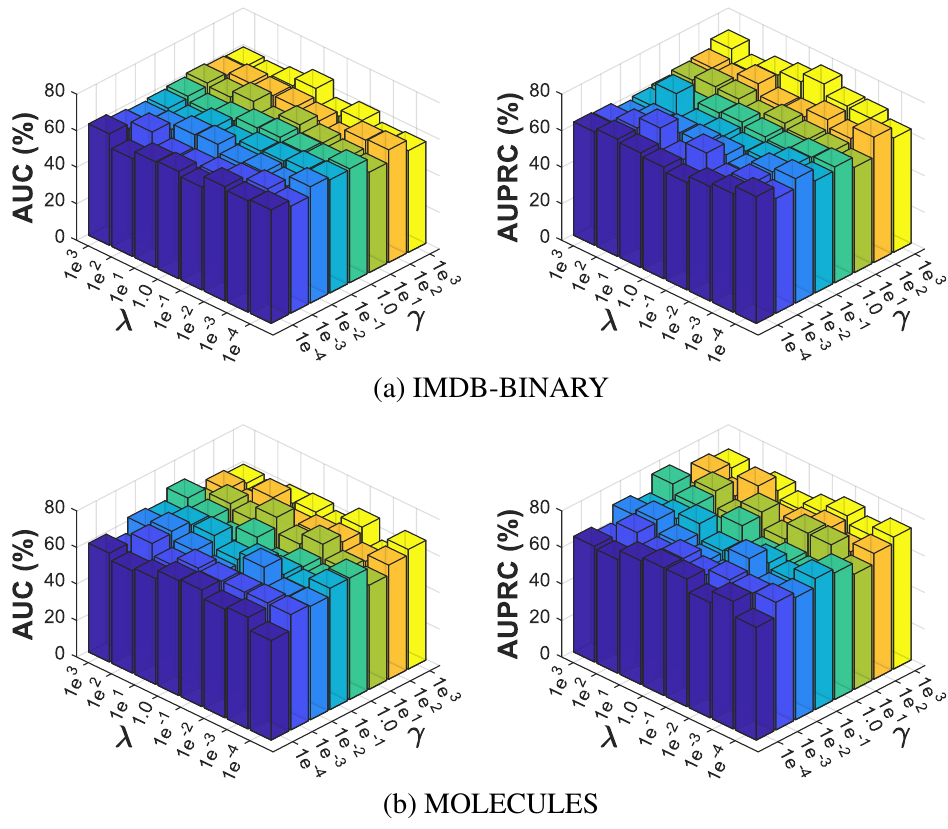}
    \caption{Parameter analysis of $\lambda$ and $\gamma$ on IMDB-BINARY and MOLECULES. Note that the values of $\lambda$ and $\gamma$ range from $[1e^{-4}, \dots, 1e^{3}]$.}
    \label{para_sensitivity}
\end{figure}

The objective function of the proposed FGAD method contains two main hyper-parameters, i.e., $\lambda$ and $\gamma$. In this section, we conduct an analysis of the impact of these two hyper-parameters on anomaly detection performance. Specifically, we vary the values of $\lambda$ and $\gamma$ within the range of $[1e^{-3}, \dots, 1e^{4}]$ and present the experimental results on IMDB-BINARY and MOLECULES datasets in Figure~\ref{para_sensitivity}. From the observations shown in the figure, we draw several conclusions. Firstly, FGAD tends to yield suboptimal performance when the values of $\lambda$ and $\gamma$ are set too low, e.g., $1e^{-4}$ and $1e^{-3}$. This emphasizes the significant role of both loss terms in the FGAD framework and suggests their effectiveness. Secondly, we can observe that excessively high values of $\lambda$ and $\gamma$ also have an adverse impact on performance, because they may obscure the primary objective of optimizing the anomaly detector. Finally, it is worth noting that FGAD exhibits relatively stable performance both in AUC and AUPRC across a wide range of $\lambda$ and $\gamma$ values, demonstrating its robustness.

\subsubsection{\textbf{Impact of Client Numbers}}
\label{sec4.5.2}
\begin{figure}[t]
    \centering
    \includegraphics[width=1\linewidth]{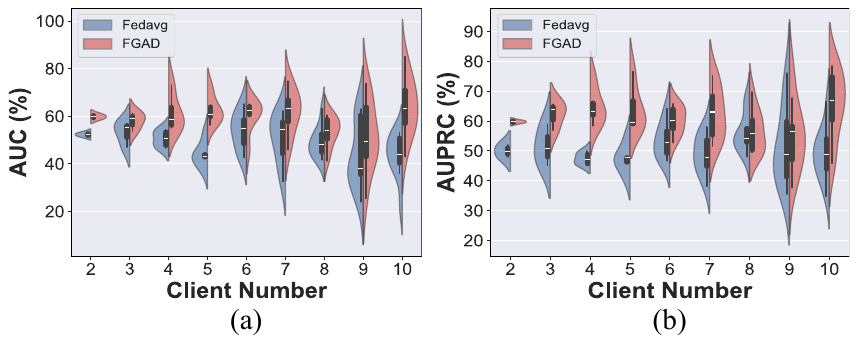}
    \caption{Average performance and distribution of variance between clients of FedAvg and FGAD. Note that the client number is set to $[2,\dots, 10]$.}
    \label{para_num_client}
\end{figure}

The number of clients $C$ is another hyper-parameter in the FGAD framework, and its impact on the performance is crucial for assessing the scalability of client numbers. Therefore, we vary the number of clients $C$ within the range of $[2, \dots, 10]$ and conduct the experiment. The results on IMDB-BINARY are reported in Figure~\ref{para_num_client}. Note that we also include FedAvg as a baseline method for comparative analysis. It can be observed that FGAD consistently achieves remarkable performance improvement compared to FedAvg in all cases, and exhibits stability against changes in the number of clients. However, when the number of clients increases to certain large values, the average performance shows a certain degradation, and the performance variance between different clients becomes more significant both in FGAD and FedAvg. This is primarily due to the gradually increasing discrepancy between the graph data distributed across different clients, which causes more severe non-IID problems. Nevertheless, FGAD still exhibits relatively smaller performance fluctuations compared with FedAvg, which fully demonstrates the scalability of the proposed FGAD method.

\subsubsection{\textbf{Impact of GIN Layers}}
\label{sec4.5.3}

\begin{figure}[t]
    \centering
\includegraphics[width=1.0\linewidth]{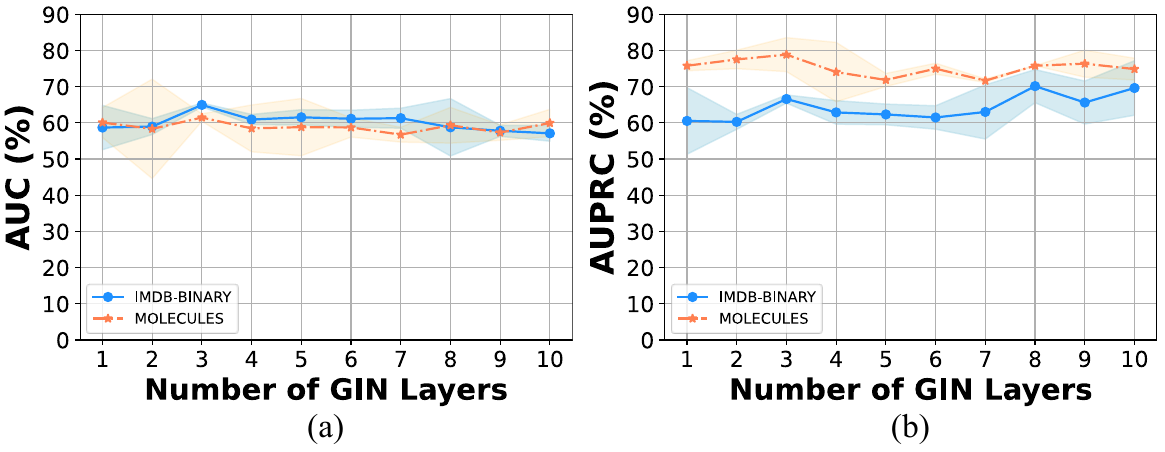}
    \caption{Average performance with standard deviation under different numbers of GIN layers on IMDB-BINARY and MOLECULES datasets. Note that the number of GIN layers is set to $[1, \dots, 10]$.}
\label{para_hidden}
\end{figure}
We delve into the impact of the number of GIN layers $K$ on the anomaly detection performance within the proposed FGAD method. The parameter $K$ plays a pivotal role in determining the extent to which the model explores neighborhood information and the overall complexity of FGAD. We systematically analyze its impact by varying the $K$ within the range of $[1, \dots, 10]$ and conduct a series of experiments. Figure~\ref{para_hidden} reports the experimental results on the IMDB-BINARY and MOLECULES datasets, from which we have the following observations. First, a certain depth of GIN is beneficial to fully leverage the structural information of graph data for learning powerful GAD models, which could be verified from the observed performance improvement. Second, when the number of GIN layers continues to increase, the observed performance improvements become increasingly marginal or even exhibit slight diminishment. This trend indicates that a moderate number of GIN layers, e.g., 3, is sufficient to effectively leverage the neighborhood information within graphs. Third, we can observe from the overall experimental results that the performance remains relatively stable under the variation of $K$, which demonstrates the robustness of FGAD.


\section{Conclusion}
\label{sec5}
In this paper, we study a challenging GAD problem with non-IID graph data distributed across multiple clients, and propose an effective federated graph anomaly detection (FGAD) method to tackle this issue. To enhance the detecting capability of local models, we propose to train a classifier in a self-boosted manner by distinguishing the normal and anomalous graphs generated from an anomaly generator. Besides, to alleviate the adverse impact of non-IID problems among clients, we introduce a student model to distill knowledge from the classifier (teacher model), and engage only the student model in collaborative learning, so that the personalization of local models could be preserved. Moreover, we improve the collaborative learning mechanism that streamlines the capacity of local models and reduces the communication costs during collaborative learning. Comprehensive experiments under various data types and scenarios compared with state-of-the-art baselines demonstrate the superiority of the proposed FGAD method. We believe that this work would pave the way for subsequent studies on collaborative GAD under the FL setting in the future.
\bibliographystyle{ACM-Reference-Format}
\bibliography{reference}

\newpage

\appendix
\begin{table*}[!ht]
\centering
\caption{Detailed information of the datasets used in the experiment.}
\setlength{\tabcolsep}{4.5mm}{
\renewcommand{\arraystretch}{1}
\begin{tabular}{lccccccc}
\toprule
Dataset Name & \#Graphs&  \#Average Nodes & \#Average Edges &\#Graph Classes & Data Type\\
\midrule
IMDB-BINARY  &1,000  &19.77  &96.53   &2 &Social Network\\
COLLAB &5,000	&74.49	&2,457.78 &3 &Social Network\\
IMDB-MULTI   &1,500	&13.00	&65.94    &3 &Social Network\\
\midrule
MUTAG        &188   &17.93  &19.79     &2 &Molecule\\
DHFR         &756	&42.43	&44.54     &2 &Molecule\\
PTC\_MR      &344   &14.29  &14.69     &2 &Molecule\\
BZR          &405   &35.75  &38.36     &2 &Molecule\\
COX2         &467   &41.22  &43.45     &2 &Molecule\\
AIDS         &2,000  &15.69  &16.20    &2 &Molecule\\
NCI1         &4,110	&29.87   &32.30    &2 &Molecule\\
\midrule
ENZYMES      &600   &32.63  &62.14   &6 &Biology\\
PROTEINS     &1,113  &39.06  &72.82   &2 &Biology\\
DD           &1,178  &284.32	&715.66  &2 &Biology\\
\bottomrule
\end{tabular}}
\label{Single_datasets}
\end{table*}

\section{Appendix}
The appendix includes the following content:
\begin{enumerate}[leftmargin=*]
\item Detailed description of graph benchmarks used in experiments.
\item Detailed experimental settings.
\item Theoretical and empirical complexity analysis.
\item Parameter analysis of latent dimensions.
\item Justification of the backbone sharing strategy.
\end{enumerate}

\subsection{Detailed Description of Graph Benchmarks}
\label{Appendix_data}
In this section, we supplement the detailed description for all the graph benchmarks used in our experiment, including the number of graphs, the average node numbers and edge numbers, and the classes. Table~\ref{Single_datasets} summarizes the information on graph benchmarks used in the experiment. Specifically, in the single-dataset experiment, we use three social network benchmarks including IMDB-BINARY, COLLAB, and IMDB-MULTI. In the multi-dataset experiment, we construct four benchmarks by integrating different types of graph data, e.g., molecules, biological, and social network data. The details are illustrated as follows:
\begin{itemize}[leftmargin=*]
\item \textbf{MOLECULES:} This benchmark includes multiple molecule datasets, e.g., MUTAG, DHFR, PTC\_MR, BZR, COX2, AIDS, and NCI1.
\item \textbf{BIOCHEM:} This benchmark is a cross-domain dataset including datasets in MOLECULE, and additional biological datasets, e.g., ENZYMES, PROTEINS, and DD.
\item \textbf{SOCIALNET:} This benchmark includes multiple social network datasets, e.g., IMDB-BINARY, COLLAB and IMDB-MULTI.
\item \textbf{MIX:} This benchmark contains all datasets from three domains, i.e., molecular, biological, and social network, in Table~\ref{Single_datasets}.
\end{itemize}
Note that we regard the graph in the first class of each dataset as the normal graph, and the graphs in other classes as anomalous graphs. All graph benchmarks used in this paper source from TUDataset~\cite{Morris2020}, a publicly available graph benchmark database\footnote{https://chrsmrrs.github.io/datasets/docs/datasets/}.

\subsection{Detailed Experimental Settings}
\label{Appendix_experimental_setting}
In this section, we supplement more details of the experimental settings in the paper, including the network structure, trade-off parameter settings, training details, baseline settings, etc.
\begin{itemize}[leftmargin=*]
    \item \textbf{Network Structure:}  We employ a 3-layer GIN~\cite{xu2019powerful} as the backbone network for our method, with the aggregated dimension in each layer set to 64. In addition, we adopt the 4-layer and 3-layer fully connected networks for the teacher head and student head, respectively. The network structure of the teacher head is set to 256-192-128-64-2, while for the student head is 192-128-64-2. Moreover, we will open-source the code of FGAD for details and reproducibility.
    \item \textbf{Data Split: } For all datasets, we regard the graphs in the first class as normal and graphs in other classes as anomalous. We allocate 80\% of the normal graphs data for training, and subsequently construct the testing data by combining the remaining normal data with an equal number of anomalous graphs.
    \item \textbf{Training Details:} We fix the batch size as 64 for all experiments and use Adam~\cite{kingma2013auto} as the optimizer with a fixed learning rate $\alpha = 0.001$. We first pre-train each local model excluding the student network and knowledge distillation module for 10 epochs. Then we jointly train the whole network with collaborative learning for 200 epochs.
    \item \textbf{Trade-off Parameter Settings:} The objective function of FGAD contains two trade-off parameters, i.e., $\lambda$, and $\gamma$, we vary their values within the range of $[1e^{-4}, 1e^{3}]$ and evaluate their impact on performance in the Section~\ref{sec4.5.1}.
    Regarding the number of clients $C$ in a single-dataset, we vary it within the range of $[2, \dots, 10]$ and evaluate its impact in Section~\ref{sec4.5.2}, while for multi-dataset, the number of clients is set to the number of its sub-datasets. Besides, for the number of GIN layers $K$, we also evaluate its impact under different values in Section~\ref{sec4.5.3}.
    \item \textbf{Baseline Settings:} For the state-of-the-art baselines including FedAvg, FedProx, GCFL, and FedStar, we integrate them with DeepSVDD~\cite{ruff2018deep} to construct the end-to-end GAD model. We also include the self-training strategy that abandons collaborative learning, as one of the baselines. Besides, we employ the same GIN backbone with FGAD to guarantee the fairness of the experiment. The objective of local models in each client is to minimize the distance from the projection of the training data in the latent space to the centroid, which is randomly initialized following the setting in DeepSVDD and fixed throughout the training phase. In the collaborative learning phase, we upload the learned decision boundaries in each client as part of the parameters and aggregate them in the server. Finally, we can calculate the anomaly score by the distances between the graph representation and the centroid after training, and the smaller the score, the more the graph tends to be considered normal.
    \item \textbf{Implementation:} The implementation of FGAD is based on PyTorch Geometric~\cite{fey2019fast} library, and the experiments are run on NVIDIA Tesla A100 GPU with AMD EPYC 7532 CPU. 
\end{itemize}

\subsection{Theoretical Complexity Analysis}
\label{Appendix_theoretical_complexity}
Here we provide theoretical complexity analysis of the proposed FGAD method. Assume there are $N$ graphs across all clients, and with maximal $m$ nodes and $|E|_{\mathrm{max}}$ edges within a graph. In the local model of each client, the maximal dimension among input and latent space of GIN is denoted by $\tilde{d}$, and the number of GIN layers is represented by $L$. In Addition, the maximal latent dimensions of the teacher and student heads are denoted by $d_{\mathrm{t}}$ and $d_{\mathrm{s}}$, respectively. Besides, the number of latent layers in the teacher and student heads is denoted by $K_{\mathrm{t}}$ and $K_{\mathrm{s}}$. Subsequently, we analyze the time and space complexity of FGAD within a single client, as well as the communication complexity in collaborative learning, as follows:
\begin{itemize}[leftmargin=*]
\item \textbf{Time Complexity}: Since the teacher and student models share the same GIN backbone, the time complexity of the backbone network is $\mathcal{O}(NL(m\tilde{d}^{2} + |E|_{\mathrm{max}}\tilde{d}))$. Similarly, the time complexity of the anomaly generator in the teacher model mainly comes from the GIN. For the teacher and student heads, the time complexities are $\mathcal{O}(K_{\mathrm{t}}\tilde{d}d_{\mathrm{t}})$ and $\mathcal{O}(K_{\mathrm{s}}\tilde{d}d_{\mathrm{s}})$, respectively. Consequently, the overall time complexity of FGAD framework is approximately $\mathcal{O}(2NL(m\tilde{d}^{2} + |E|_{\mathrm{max}}\tilde{d})+ (K_{\mathrm{t}}d_{\mathrm{t}} + K_{\mathrm{s}}d_{\mathrm{s}}) \tilde{d})$, where includes the anomaly generator weight-shared GIN backbone, and the teacher and student heads.
\item \textbf{Space Complexity}: For the space complexity of the GIN backbone, the space complexity mainly comes from the storage of weight and bias matrices in each layer, which can be denoted by $\mathcal{O}(L\tilde{d}(1+\tilde{d})$. For the teacher and student heads, their space complexities can be derived similarly, i.e., $\mathcal{O}(K_{\mathrm{t}}\tilde{d}(1+d_{\mathrm{t}}) + K_{\mathrm{s}}\tilde{d}(1+d_{\mathrm{s}}))$. Consequently, the overall space complexity of FGAD framework is approximately $\mathcal{O}(L\tilde{d}(1+\tilde{d}) + K_{\mathrm{t}}\tilde{d}(1+d_{\mathrm{t}}) + K_{\mathrm{s}}\tilde{d}(1+d_{\mathrm{s}}))$.
\item \textbf{Communication Complexity}: Since the teacher model in FGAD is used for the personalization of local clients, only the student head engages in collaboration. Consequently, the time and space complexities in a communication round are approximately $\mathcal{O}( K_{\mathrm{s}}\tilde{d}d_{\mathrm{s}})$ and $\mathcal{O}(K_{\mathrm{s}} \tilde{d}(1+d_{\mathrm{s}}))$.
\end{itemize}

\subsection{Empirical Complexity Analysis}
\label{Appendix_empirical_complexity}
\begin{figure}[t]
    \centering
    \includegraphics[width=1\linewidth]{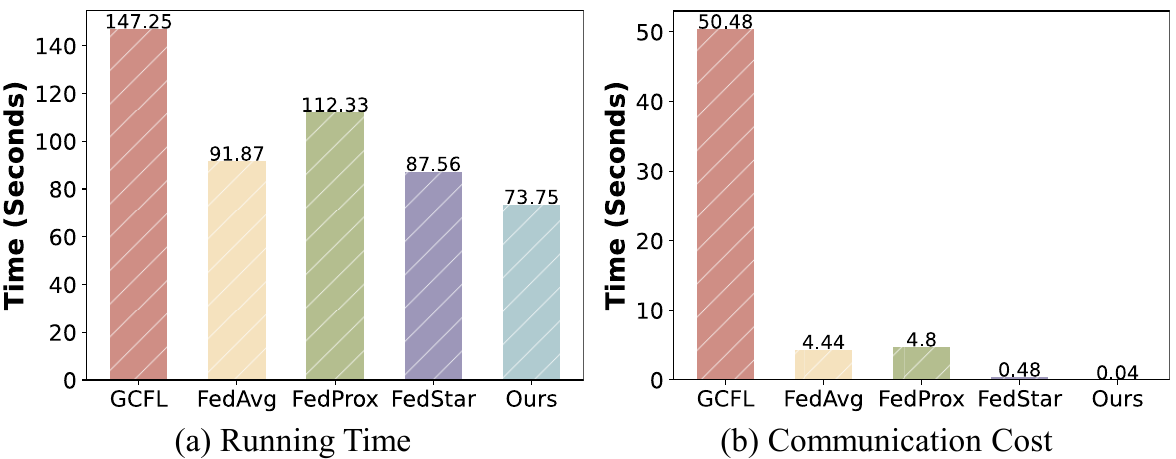}
    \caption{Running time and communication cost comparison in 200 epochs.}
    \label{time_comparison}
\end{figure}

To more comprehensively analyze the complexity of FGAD, we further provide empirical complexity analysis. Specifically, we compare the running time (in local) and communication time (in collaboration) of FGAD with other baselines. Note that the experiment is conducted under uniform device settings (detailed in Appendix A.2) to ensure fairness. The experimental results are presented in Fig. \ref{time_comparison}.

It can be observed that the time complexity of FGAD is competitive with several baselines, e.g., that of FedStar, and significantly better than that of GCFL. Combined with the performance comparison in Tables~\ref{result} and \ref{result2} (in the paper), the overall experimental results demonstrate that FGAD not only significantly improves anomaly detection performance but also possesses promising time efficiency compared to other baselines.

Additionally, communication cost (time) is also an important evaluation metric in federated learning. Therefore, we further conduct the comparative experiment to demonstrate the effectiveness of FGAD. As shown in Fig. \ref{time_comparison} (b), FGAD has the lowest communication time compared with other baselines, which aligns with the comparison of exchanging amount of network parameters in Table~\ref{result}. It should be noted that this is the analog communication time without considering the network bandwidth. When it comes to real-world collaboration, the network bandwidth will significantly impact the efficiency of model parameter transmission. Consequently, in cases of models with large parameter sizes, the communication time becomes a pivotal factor influencing the time complexity of collaborative learning.

\subsection{Impact of Latent Dimensions}
\label{Appendix_latent_dimension}
\begin{figure}[t]
    \centering
    \includegraphics[width=1\linewidth]{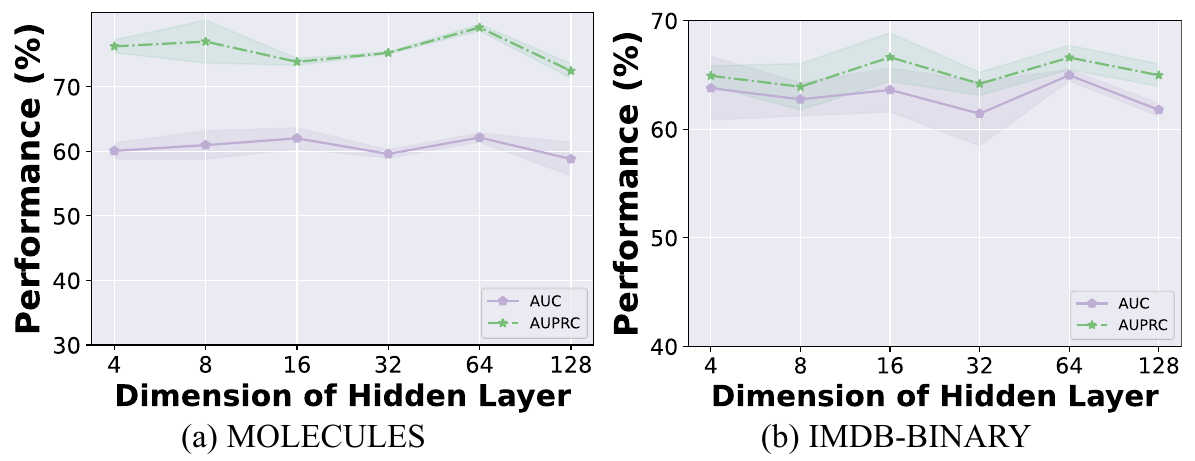}
    \caption{Parameter sensitivity of different dimensions for hidden layers.}
    \label{parameter_hidden_dim}
\end{figure}

Here, we further conduct additional parameter analysis for the impact of the latent dimension in the GIN layer. Specifically, we set the latent dimension from $[4,128]$, and the experimental results on MOLECULES and IMDB-BINARY shown in Fig. \ref{parameter_hidden_dim}.
The results suggest that FGAD exhibits relatively stable performance across a wide range of latent layer dimensions, demonstrating its robustness. Nevertheless, it can be observed that excessively high dimensions (e.g., 128) might adversely affect performance, potentially due to the redundant information it brings.

\subsection{Justification of the Backbone Sharing}
\label{Appendix_backbone_share}
To justify the rationale for sharing the backbone network between the teacher and student models, we conduct additional experiments by comparing the performance of FGAD with and without sharing the GIN backbone. The results are presented in Table~\ref{Backbone_share}. We can observe only a marginal difference in performance between these two strategies. This observation suggests that sharing the GIN backbone would not decrease the effectiveness of knowledge distillation in FGAD. More importantly, the significant benefit of sharing the GIN backbone is the substantial reduction in model complexity. This streamlined architecture leads to a more efficient model in terms of computational resources and memory usage.
\begin{table}[h]
\centering
\caption{Peformance (mean(\%) $\pm$ std(\%)) of FGAD under shared/unshared GIN backbone.}
\setlength{\tabcolsep}{0.8mm}{
\renewcommand{\arraystretch}{1}
\begin{tabular}{l||c|c|c|c}
\toprule
\multirow{2}{*}{Backbone} & \multicolumn{2}{c|}{IMDB-BINARY} &\multicolumn{2}{c}{IMDB-MULTI} \\ \cmidrule{2-5}
& AUC & AUPRC & AUC & AUPRC\\ \midrule
Shared GIN &64.97$\pm$0.52&66.60$\pm$1.12&60.51$\pm$1.18&66.82$\pm$0.14 \\
w/o Shared GIN  &63.13$\pm$1.19 &66.43$\pm$2.23&58.13$\pm$0.84&66.67$\pm$0.00 \\
\bottomrule
\end{tabular}}
\label{Backbone_share}
\end{table}

\end{document}